\documentclass[sigconf]{acmart}

\AtBeginDocument{%
  \providecommand\BibTeX{{%
    \normalfont B\kern-0.5em{\scshape i\kern-0.25em b}\kern-0.8em\TeX}}}

\copyrightyear{2021}
\acmYear{2021}
\setcopyright{acmlicensed}\acmConference[IH\&MMSec '21]{Proceedings of the 2021 ACM Workshop on Information Hiding and Multimedia Security}{June 22--25, 2021}{Virtual Event, Belgium}
\acmBooktitle{Proceedings of the 2021 ACM Workshop on Information Hiding and Multimedia Security (IH\&MMSec '21), June 22--25, 2021, Virtual Event, Belgium}
\acmPrice{15.00}
\acmDOI{10.1145/3437880.3460411}
\acmISBN{978-1-4503-8295-3/21/06}

\usepackage{tikz}
\usetikzlibrary{intersections}
\tikzset{>=stealth}
\tikzset{point/.style={draw,circle,uibkblue,inner sep=.8pt}}

\providecommand{\confusionCell}[2]{
    \node[rectangle, minimum size=4mm, fill=uibkblue!#2, align=center, draw=lightgray] at (\x, \y) {\color{\ifnum#2>40 white\else black\fi}\scriptsize#1};
    \pgfmathparse{\x+0.4}
    \edef\x{\pgfmathresult}
}
\providecommand{\confusionNewline}{
    \edef\x{0}
    \pgfmathparse{\y-0.4}
    \edef\y{\pgfmathresult}
}

\definecolor{uibkblue}{cmyk}{1,0.6,0,0.65}%
\definecolor{uibkorange}{cmyk}{0,.5,1,0}%
\definecolor{uibkgray}{cmyk}{0,0,0,0.6}%
\definecolor{red}{rgb}{0.8,0.06,0.00}%

\tikzstyle{color1} = [color=red]
\tikzstyle{color2} = [color=uibkorange]
\tikzstyle{color3} = [color=uibkblue]

\usepackage{amsmath}

\DeclareMathOperator{\sign}{sign}
\providecommand{\x}{\ensuremath{\mathbf{x}}}
\providecommand{\dconf}{\ensuremath{\delta_{\textup{conf}}}}

\usepackage{subcaption}
\usepackage{algorithm}
\usepackage[noend]{algpseudocode}
\usepackage{siunitx}
\usepackage{cleveref}
\usepackage{hyperref}
\usepackage{color}
\usepackage{booktabs}
\usepackage{multirow}
\usepackage{bm}

\usetikzlibrary{calc}

\providecommand{\Aq}{\vphantom{Aq}}
\providecommand{\MA}{MA}
\providecommand{\MAs}{MAs}

\providecommand{\NNs}{NNs}
\providecommand{\local}{local}
\providecommand{\remote}{remote}

\begin{document}

\title{i{NN}formant: Boundary Samples as Telltale Watermarks}

\author{Alexander Schlögl}
\email{alexander.schloegl@uibk.ac.at}
\affiliation{%
  \institution{Department of Computer Science}
  \institution{University of Innsbruck}
  \country{Austria}
}

\author{Tobias Kupek}
\email{tobias.kupek@swarm-analytics.com}
\affiliation{%
  \institution{Swarm Analytics GmbH}
  \country{Austria}
}
\authornote{Work carried out while at the University of Innsbruck.}

\author{Rainer Böhme}
\email{rainer.boehme@uibk.ac.at}
\affiliation{%
  \institution{Department of Computer Science}
  \institution{University of Innsbruck}
  \country{Austria}
}





\begin{abstract}
  Boundary samples are special inputs to artificial neural networks crafted to identify the execution environment used for inference by the resulting output label.
  The paper presents and evaluates algorithms to generate transparent boundary samples.
  Transparency refers to a small perceptual distortion of the host signal (i.\,e., a natural input sample).
  For two established image classifiers, ResNet on FMNIST and CIFAR10, we show that it is possible to generate sets of boundary samples which can identify any of four tested microarchitectures.
  These sets can be built to not contain any sample with a worse peak signal-to-noise ratio than \SI{70}{dB}.
  We analyze the relationship between search complexity and resulting transparency.\footnote{Code can be found at \url{https://github.com/alxshine/innformant}}
\end{abstract}




\maketitle

\section{Introduction}
Recently it has been observed that the numerical predictions of neural networks (\NNs) vary between different CPU microarchitectures (\MAs{})~\cite{schloegl2021forensicability}.
This can be used in forensic investigations to identify the execution environment used for predictions, or verify that a prediction has been made on specific hardware.

These numerical differences may also offer novel ways to implement digital rights management for trained machine learning models.
For instance, the owner of a model could verify if a given prediction has been generated on licensed hardware, which might be equipped with a secure billing device.
Predictions whose numerical values indicate the use of \emph{another} \MA{} indicate that the billing mechanism might have been bypassed fraudulently.

A major obstacle to this application is that the numerical differences between \MAs{} are tiny.
They almost always disappear at the last step of the inference pipeline when a real-valued soft-max vector is quantized to a symbolical label.
Boundary samples fix this problem.
These samples lie in the area between decision boundaries that arise from the numerical differences between MAs, as was observed in \cite{schloegl2021forensicability}.
Boundary samples are then classified differently depending on the execution environment, allowing the owner of a model in the above example to probe the hardware used for prediction.
In the best case, any deviation from the licensed hardware's \MAs{} is detectable by the class label only.

Boundary samples are barely researched.
It may appear surprising even that they exist and can be found efficiently as claimed in~\cite{schloegl2021forensicability}.
This work adds another consideration, namely the transparency of boundary samples.
This is relevant if, in the above example, the model owner wants to probe the inference pipeline inconspicuously in order to avoid that the licensee can process obvious boundary samples in a different pipeline (the legitimate one) than the bulk of organic samples.
We propose to generate transparent boundary samples as perturbations of natural input samples and measure the distortion by the peak signal-to-noise ratio (PSNR).

\begin{figure}[h]
  \begin{tikzpicture}
    \draw
    (-3,2) rectangle (3,-2);

    \path
    (-1.5,2) coordinate (top0)
    (.9,-2) coordinate (bottom0)
    (-1,2) coordinate (top1)
    (1.1,-2) coordinate (bottom1)
    ;

    \begin{scope}[fill=uibkorange]
        \fill (top0) -- (top1) -- (bottom1) -- (bottom0) -- cycle;
    \end{scope}

    \begin{scope}[fill=uibkblue!30]
        \fill (top1) -- (bottom1) -| (3,2) |- cycle;
    \end{scope}

    \draw[thick]
    (bottom0) -- (top0) node[pos=.8,left] {$b_0$}
    (bottom1) -- (top1) node[pos=.8,right] {$b_1$};

    \begin{scope}[xshift=-2.5cm,yshift=-.8cm]
        \scriptsize
        \draw
        (0,0) rectangle +(2.2,-.6);

        \draw [fill=uibkorange]
        ++(.1,-.1) rectangle +(.25,-.15) node[midway,right=3pt] {Boundary Samples\Aq};

        \draw [fill=uibkblue!30]
        ++(.1,-.35) rectangle +(.25, -.15) node[midway,right=3pt] {Adversarial Samples\Aq};
    \end{scope}
\end{tikzpicture}
  \caption{Input space for adversarial and boundary samples, for two given decision boundaries $b_0$ and $b_1$.}
  \label{fig:difficulty}
\end{figure}
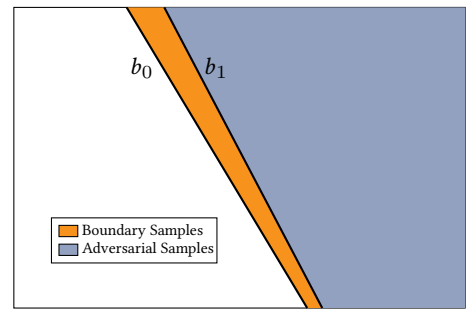

There are striking parallels to digital watermarking~\cite{cox2007digitalwatermarking}:
the natural input sample is the host signal and the perturbation is a \emph{telltale watermark}, i.\,e., a special case of fragile watermark designed to indicate the type of processing applied to the watermarked signal~\cite{kundur1999telltale,carnein2016telltalejpg}.
Note that we do not aim for undetectability in the sense of secure steganography.
The facts that boundary samples are rare, and good ones are classified differently by any two \MAs{}, seem to preclude any attempt to make them undetectable for anyone who can run the model on multiple \MAs{}.

There are also parallels to evasion attacks in adversarial machine learning~\cite{papernot2018securityML}.
\Cref{fig:difficulty}~gives some intuition on the difficulty of finding boundary samples compared to the problem of finding adversarial samples.
Both problems have in common that an input in the white region has to be perturbed to fall into another region subject to small perceptual distortion.
While the solution space for adversarial samples is the entire blue area, boundary samples must hit the orange region.
In fact, adversarial samples try to move as ``deep'' as possible (given the perturbation constraints) into the blue area in order to be transferable~\cite{Szegedy2014Intriguing}.
In contrast, boundary samples must hit the small orange area between the decision boundaries that arises purely from numerical differences between \MAs{}.

This short paper is organized as follows.
The next section presents our generation method for boundary samples, which is a modification of FGSM, a known search algorithm for generating adversarial samples~\cite{goodfellow2014Harnessing}.
Section~\ref{sec:eval} evaluates the effectiveness and efficiency of the proposed algorithm on two standard pairs of dataset and model architectures (FMNIST with ResNet20 and CIFAR10 with ResNet32).
We report runtime measurements (in terms of iterations broken down by \MAs), the resulting success rates, and distortions.
Section~\ref{sec:related} discusses related work.
The concluding Section~\ref{sec:future} points out limitations and shows directions for future work.


\section{Generating Boundary Samples}

Our method makes the following assumptions.
We have white-box access to a trained feed-forward deep neural network and oracle access to predictions and gradients from that network on a closed set of relevant \MAs{}.
Access costs vary between oracles.
We assume that one \MA{} is \emph{local} and gives us a cheap (fast) oracle.
All other \MAs{} are \emph{remote} and may in practice be more costly (slower).
Moreover, we require access to a number of test samples drawn from the training distribution.
Transparency is defined by the distortion between a given test sample (the starting point for the iterative algorithms) and the resulting boundary sample.
We consider a set of boundary samples as \emph{fully identifying} if it contains at least one sample that is predicted with a unique label for each \MA{} in the set.


We proceed in two steps.
In the next subsection, we examine the case of a binary decision problem between two candidate \MAs{}.
Then, in \Cref{ssec:1vn}, we generalize to the case of identifying one out of $n$ candidate \MAs{}.
This is still a binary decision problem, but the solution space is much more constrained.

\subsection{The 1 vs 1 Case}
\label{ssec:1v1}

The problem of differentiating between two candidate \MAs{} is equivalent to constructing a sample that falls in the orange regions in \Cref{fig:difficulty}.
We split the generation into a \emph{local} and a \emph{remote} phase.

\paragraph{Local phase} In the local phase we try to get as close to the decision boundary as possible.
We do this with the modified iterative fast-gradient-sign method (i-FGSM)~\cite{goodfellow2014Harnessing}.
FGSM was chosen based on the intuition that many small perturbations lead to a lower mean square error (and hence higher PSNR) than fewer larger perturbations.
For model $m$, input $\x$, and a step size $\alpha$, the $i$-th FGSM step works as follows,
\begin{equation}
  \x_i = \x_{i-1} + \alpha \sign\left(\nabla_{\x} m(\x_{i-1})\right).
\end{equation}

Compared to FGSM our algorithm flips the sign based on the correctness of the prediction, and reduces the step size as we approach the decision boundary.
Varying the perturbation's sign lets us approach the decision boundary even after overshooting.
As we approach the decision boundary, the confidence difference $\dconf$ between the first and second predicted classes decreases.
While the gradients' norms could be used to judge the distance to the decision boundary, we used $\dconf$ as an approximate distance measure.
Reducing the step size along with $\dconf$ allows us to gradually approach the decision boundary, until a termination condition is met.
The modified FGSM step looks as follows,
\begin{equation}
  \x_i = \x_{i-1} + c\ \dconf\ \alpha \sign\left(\nabla_\x m\left(\x_{i-1}\right)\right),
  \label{eq:modified}
\end{equation}
where $c$ is the correctness sign.
It takes value $1$ if $\x_{i-1}$ is misclassified, and $-1$ otherwise.

As local predictions are cheaper than remote predictions, we want to approach the decision boundary as close as possible with local steps; specifically to a confidence difference of less than $10^{-6}$, or ideally $10^{-7}$.
Choosing the right $\alpha$ is crucial.
If it is too high, the sample bounces around the decision boundary without approaching it.
If it is too low, the sample movement stalls as the confidence difference vanishes.
This can be detected if the predictions are identical in two consecutive steps.
In this case, we multiply $\dconf$ with a scaling factor.
We increase $\dconf$ exponentially in this fashion until the predictions change again, at which point $\dconf$ is reset to the new confidence difference.
We use $10^{-4}$ as value for $\alpha$.

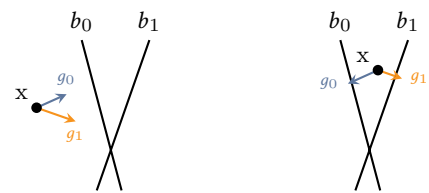
\begin{figure}[h]
  \begin{subfigure}{.45\columnwidth}
\centering
\begin{tikzpicture}
    \path
    (-.5,1) coordinate (top1)
    (.03,-1) coordinate (bottom1)
    (.4,1) coordinate (top2)
    (-.3,-1) coordinate (bottom2);

    \draw[thick]
    (bottom1) -- (top1) node[above] {$b_0$}
    (bottom2) -- (top2) node[above] {$b_1$};

\coordinate (x) at (-1.1,.1);

\begin{scope}[->,thick,scale=.2]
\draw[uibkblue!50]
(x) -- ++(2,.9) node[above] {\scriptsize $g_0$}; 
\end{scope}
\begin{scope}[->,thick,scale=.26]
\draw[uibkorange]
(x) -- ++(2,-.7) node[below] {\scriptsize $g_1$};
\end{scope}
\fill (x) circle (.07) node[above left] {\x} ;
\end{tikzpicture}
\caption{Same class predicted}
\end{subfigure}%
\begin{subfigure}{.45\columnwidth}
\centering
\begin{tikzpicture}
    \path
    (-.5,1) coordinate (top1)
    (.03,-1) coordinate (bottom1)
    (.4,1) coordinate (top2)
    (-.3,-1) coordinate (bottom2);

    \draw[thick]
    (bottom1) -- (top1) node[above] {$b_0$}
    (bottom2) -- (top2) node[above] {$b_1$};

\coordinate (x) at (-.0,.6);

\begin{scope}[->,thick,scale=.2]
\draw[uibkblue!50]
(x) -- ++(-2,-.9) node[left] {\scriptsize $g_0$}; 
\end{scope}
\begin{scope}[->,thick,scale=.16]
\draw[uibkorange]
(x) -- ++(2,-.7) node[right] {\scriptsize $g_1$};
\end{scope}
\fill (x) circle (.07) node[above left] {\x} ;
\end{tikzpicture}
\caption{Different class predicted}
\end{subfigure}
  \caption{Possible cases for boundary sample predictions.}
  \label{fig:boundaryCases}
\end{figure}

\paragraph{Remote phase}  Once we are sufficiently close with the local oracle, we use gradients from our remote oracles to further refine the boundary sample.
In this remote phase, one of three cases occurs:
\begin{enumerate}
  \item The label flips on neither instance.
  \item The label flips on one instance, but not the other.
  \item The label flips on both instances.
\end{enumerate}
In the second case, we have found a boundary sample and terminate.
Note that the third case is symmetric to the first, and our handling is identical.
The possible cases are shown in \Cref{fig:boundaryCases}, where $g_i$ denotes the gradient for instance $i$.

When both instances predict the same class, we further approach the closest decision boundary using the same modified FGSM step as in the local phase.
Our remote step requests predictions and gradients from both oracles, and then uses the one where the corresponding $\dconf$ is smaller for the FGSM step shown in \Cref{eq:modified}.
If the labels from both instances flip together, we simply continue as $c$ will correct the direction of our perturbation even if the closest boundary is not the same as before.
This process is repeated until either a successful boundary sample is generated or a maximum number of iterations is reached.
The full algorithm is shown in \Cref{alg:1vN}, where the lines relevant to the 1 vs 1 case are highlighted.

\paragraph{Remarks} The local and remote phases are very similar.
In principle, one could also omit the local preparation entirely and start with a clean sample in the remote phase.
As we used cloud instances for our remote oracles, remote predictions were slower and much more costly than the local predictions.

\subsection{The 1 vs $n$ Case}
\label{ssec:1vn}
\begin{figure}[h]
  \begin{tikzpicture}[y=6.5mm]
    \providecommand{\intersect}[2]{\coordinate (i#1#2) at (intersection of top#1--bottom#1 and top#2--bottom#2);}

    \path
    (0,1.5) coordinate (top0)
    (-0.1,-1.5) coordinate (bottom0)
    (.4,1.5) coordinate (top1)
    (.5,-1.5) coordinate (bottom1)
    (.8,1.5) coordinate (top2)
    (.2,-1.5) coordinate (bottom2)
    (1.3,1.5) coordinate (top3)
    (.6,-1.5) coordinate (bottom3)
    ;
    \providecommand{\n}{3}

    \intersect{1}{2}

    \begin{scope}[fill=uibkorange]
        \fill (top0) -- (bottom0) -- (bottom2) -- (i12) -- (top1) -- cycle;
    \end{scope}

    \begin{scope}[fill=uibkblue!30]
        \fill (top3) -- (top2) -- (i12) -- (bottom1) -- (bottom3) -- cycle;
    \end{scope}

    \scriptsize
    \foreach \i in {0,...,\n}
    \draw[thick] (bottom\i) -- (top\i) node[above] {$b_\i$};


    \begin{scope}[xshift=2.4cm,yshift=0.8cm]
        \normalsize
        \draw (0,0) rectangle +(8.3em,-10.5ex);

        \draw[fill=uibkorange]
        ++(.5em,-1ex) rectangle +(1em,-2.5ex) node[midway,right=3pt] {identifies $A_0$\Aq};
        \draw[fill=uibkblue!30]
        ++(.5em,-4ex) rectangle +(1em,-2.5ex) node[midway,right=3pt] {identifies $A_3$\Aq};
        \draw
        ++(.5em,-7ex) rectangle +(1em,-2.5ex) node[midway,right=3pt] {not identifying\Aq};
    \end{scope}

    \begin{scope}
        \normalsize
        \draw (-.8,-.3) node {class $u$};
        \draw (1.5,-.3) node {class $v$};
    \end{scope}
\end{tikzpicture}
  \caption{Uniquely identifying areas for multiple instances.}
  \label{fig:identifying}
\end{figure}
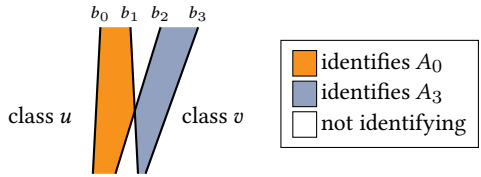

We can use the same approach as in \Cref{fig:boundaryCases} to identify a single \MA{} from a set of $n$ known \MAs{}.
The only relevant modification concerns the selection of the target gradient.
The choice of target boundary is given by the requirement of a 1 vs $n$ boundary sample: one \MA{} results in a different label compared to all others.
This requires our boundary sample to lie past the decision boundary for one \MA{}, but before the decision boundary for all others, or vice versa.
\Cref{fig:identifying} highlights areas around the decision boundaries that uniquely identify MAs.

The 1 vs $n$ boundary sample generation process discussed here is not targeted.
This means we cannot choose the \MA{} to be identified, but let the algorithm find any \MA{} that can be identified from all others in the set.
This is not a big limitation as we shall see in Section~\ref{sec:eval} that different starting samples let us identify different \MAs{}, and each \MA{} is singled out sufficiently often.
Hence, we can repeatedly run the algorithm until we find a boundary sample which identifies any desired \MA{}.

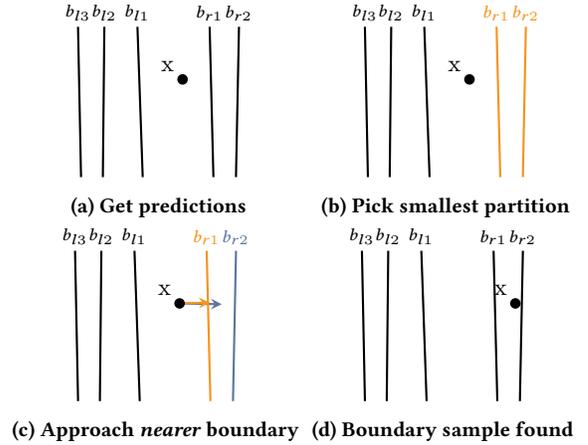
\begin{figure}[h]
  \begin{subfigure}{.45\columnwidth}
    \centering
    \begin{tikzpicture}
        \path
        (-.5,1) coordinate (lt1)
        (-.43,-1) coordinate (lb1)
        (-.94,1) coordinate (lt2)
        (-.96,-1) coordinate (lb2)
        (-1.29,1) coordinate (lt3)
        (-1.25,-1) coordinate (lb3);

        \path
        (.46,1) coordinate (rt1)
        (.51,-1) coordinate (rb1)
        (.85,1) coordinate (rt2)
        (.81,-1) coordinate (rb2);

        \begin{scope}[thick]

            \scriptsize
            \foreach \i in {1,...,3}
            \draw (lb\i) -- (lt\i) node[above] {$b_{l\i}$};

            \foreach \i in {1,2}
            \draw (rb\i) -- (rt\i) node[above] {$b_{r\i}$};
        \end{scope}

        \fill (.1,.3) circle (.07) node[above left] {$\x$};
    \end{tikzpicture}
    \caption{Get predictions}
    \label{fig:1vN-request}
\end{subfigure}%
\begin{subfigure}{.45\columnwidth}
    \centering
    \begin{tikzpicture}
        \path
        (-.5,1) coordinate (lt1)
        (-.43,-1) coordinate (lb1)
        (-.94,1) coordinate (lt2)
        (-.96,-1) coordinate (lb2)
        (-1.29,1) coordinate (lt3)
        (-1.25,-1) coordinate (lb3);

        \path
        (.46,1) coordinate (rt1)
        (.51,-1) coordinate (rb1)
        (.85,1) coordinate (rt2)
        (.81,-1) coordinate (rb2);

        \begin{scope}[thick]

            \scriptsize
            \foreach \i in {1,...,3}
            \draw (lb\i) -- (lt\i) node[above] {$b_{l\i}$};

            \begin{scope}[uibkorange]
                \foreach \i in {1,2}
                \draw (rb\i) -- (rt\i) node[above] {$b_{r\i}$};
            \end{scope}
        \end{scope}

        \fill (.1,.3) circle (.07) node[above left] {$\x$};
    \end{tikzpicture}
    \caption{Pick smallest partition}
    \label{fig:1vN-partition}
\end{subfigure}

\begin{subfigure}{.45\columnwidth}
    \centering
    \begin{tikzpicture}
        \path
        (-.5,1) coordinate (lt1)
        (-.43,-1) coordinate (lb1)
        (-.94,1) coordinate (lt2)
        (-.96,-1) coordinate (lb2)
        (-1.29,1) coordinate (lt3)
        (-1.25,-1) coordinate (lb3);

        \path
        (.46,1) coordinate (rt1)
        (.51,-1) coordinate (rb1)
        (.85,1) coordinate (rt2)
        (.81,-1) coordinate (rb2);

        \path
        (.1,.3) coordinate (x);

        \begin{scope}[thick]

            \scriptsize
            \foreach \i in {1,...,3}
            \draw (lb\i) -- (lt\i) node[above] {$b_{l\i}$};

            \begin{scope}[uibkblue!50]
                \draw (rb2) -- (rt2) node[above] {$b_{r2}$};
                \begin{scope}[scale=.28]
                    \draw[->] (x) -- ++(2,-.04);
                \end{scope}
            \end{scope}

            \begin{scope}[uibkorange]
                \draw (rb1) -- (rt1) node[above] {$b_{r1}$};
                \begin{scope}[scale=.2]
                    \draw[->] (x) -- ++(2,.05);
                \end{scope}
            \end{scope}
        \end{scope}

        \fill (.1,.3) circle (.07) node[above left] {$\x$};
    \end{tikzpicture}
    \caption{Approach \emph{nearer} boundary}
    \label{fig:1vN-approach}
\end{subfigure}%
\begin{subfigure}{.45\columnwidth}
    \centering
    \begin{tikzpicture}
        \path
        (-.5,1) coordinate (lt1)
        (-.43,-1) coordinate (lb1)
        (-.94,1) coordinate (lt2)
        (-.96,-1) coordinate (lb2)
        (-1.29,1) coordinate (lt3)
        (-1.25,-1) coordinate (lb3);

        \path
        (.46,1) coordinate (rt1)
        (.51,-1) coordinate (rb1)
        (.85,1) coordinate (rt2)
        (.81,-1) coordinate (rb2);

        \begin{scope}[thick]

            \scriptsize
            \foreach \i in {1,...,3}
            \draw (lb\i) -- (lt\i) node[above] {$b_{l\i}$};

            \foreach \i in {1,2}
            \draw (rb\i) -- (rt\i) node[above] {$b_{r\i}$};
        \end{scope}

        \fill (.75,.3) circle (.07) node[above left] {$\x$};
    \end{tikzpicture}
    \caption{Boundary sample found}
    \label{fig:1vN-found}
\end{subfigure}
  \caption{Steps for generating 1 vs $n$ boundary samples.}
  \label{fig:1vN}
\end{figure}

The algorithm proceeds as follows.
As before, we first approach the nearest local decision boundary as close as possible.
Starting with the local oracle ensures that the farthest distance is traversed with cheap (fast) queries.
We then request predictions from the remote oracles, which tell us where our current sample lies with regard to all \MAs{}' decision boundaries.
This step is shown in \Cref{fig:1vN-request}, where the decision boundaries are indexed with $c$ and $r$ for ``left'' and ``right'' for convenience.
We partition all predictions according to their label and select the smallest partition (\Cref{fig:1vN-partition}).
From the smallest partition, we choose the \emph{second farthest} decision boundary, i.\,e., the one with the second highest $\dconf$, as our target (\Cref{fig:1vN-approach}).
Passing the second furthest decision boundary leaves only one label flipped from all others (\Cref{fig:1vN-found}), meaning the generated boundary sample uniquely identifies an \MA{} (the rightmost \MA{} in the example).
\Cref{fig:1vN} illustrates our algorithm, and \Cref{alg:1vN} gives the pseudocode.

\paragraph{Remarks}
We optimize our target selection for cases when all \MAs{} return the same label, in which case we approach the closest decision boundary.
This happens in \cref{alg:closest}.
Moreover, in our experiments we had two oracles each for several \MAs{}.
We thus had to modify the exit condition to not only exit if a single label is different from all others.
We also checked whether the labels from an entire \MA{} are different from all others, but identical to each other.
This special case is included in \cref{alg:exit}.

\begin{algorithm}
  \caption{Untargeted boundary sample generation.}
  \label{alg:1vN}

  \begin{algorithmic}[1]
    \newif\ifboldnumber
    \newcommand{\boldnext}{\global\boldnumbertrue}

    \algrenewcommand\alglinenumber[1]{%
        \footnotesize\ifboldnumber\bfseries\textcolor{uibkorange}{#1:}\else#1:\fi\global\boldnumberfalse}

    \boldnext
    \Procedure{GenerateBoundarySample}{$\x,\textit{servers}$}
    \boldnext
    \For{$i \gets 1,...,\textit{local\_max}$}
    \Comment{local phase}
    \boldnext
    \State $\textup{predict } \x \textup { locally}$
    \boldnext
    \If {$\dconf < \textit{target}$}
    $\textbf{break}$
    \EndIf
    \boldnext
    \State $\x \gets \x + l\ \dconf\ \alpha\ \sign\left(\nabla_\x m\left(\x\right)\right)$
    \EndFor

    \boldnext
    \For{$i \gets 1,...,\textit{remote\_max}$}
    \Comment{remote phase}
    \boldnext
    \State $\textit{results} \gets \textup{predict } \x \textup{ on each }\textit{server}$
    \State $\textit{groups} \gets \textit{results}\textup{ grouped by the predicted label }l$
    \State $g \gets \textup{smallest group in }\textit{groups}$
    \If {$g \textup{ contains a single \MA}$} \label{alg:exit}
    \Return x
    \EndIf
    \If {$\textup{size}\left(g\right) = n$}
    \boldnext
    \State $r \gets \textit{result} \in g \textup{ with smallest }\dconf$ \label{alg:closest}
    \Else
    \State $r \gets \textit{result} \in g \textup{ with second highest }\dconf$
    \EndIf
    \boldnext
    \State $\x \gets \x + r.l\ r.\dconf\ \alpha\ \sign\left(\nabla_\x m\left(\x\right)\right)$
    \EndFor
    \EndProcedure
\end{algorithmic}
\end{algorithm}


\section{Experimental Evaluation}
\label{sec:eval}

\paragraph{Setup}
In our experiments we set $n=4$.
\Cref{tbl:architectures} lists the \MAs{} used in all our experiments.
These are all \MAs{} available as Google Cloud instances at the time of writing, and the grouping is based on successful identifications reported in prior work~\cite{schloegl2021forensicability}.
For our evaluation we will focus on the 1 vs. $n$ case as it is the more challenging one.
We chose image classification as task and selected two common datasets of different complexity: FMNIST~\cite{xiao2017fmnist} and CIFAR10~\cite{krizhevsky2009CIFAR}.
Both datasets contain images of 10 classes.
FMNIST has an input dimension of $28\times28\times1$.
CIFAR10 has an input dimension of $32\times32\times3$.
We employ the established ResNet architecture~\cite{He2016ResNet} of two depths, namely 20 layers for FMNIST and 32 layers for CIFAR10, trained for 150 epochs each.
The Keras interface of TensorFlow version~2.3.0 was used on all \MAs{} to run predictions.
We set up a Docker container based on \texttt{tensorflow/tensorflow:2.3.0} to ensure consistency of every component above the operating system.
Using \Cref{alg:1vN}, we attempted to generate boundary samples from 400 randomly selected test images per dataset.
The termination conditions were set to 2000 for $\textit{local\_max}$ and 500 for $\textit{remote\_max}$.
We could confirm that all processes were deterministic on each \MA{} (but certainly not across them).

\begin{table}
  \caption{Overview of the architectures used in this work.}
  \label{tbl:architectures}
  \begin{tabular}{ll}
    \toprule
    \textbf{Label} & \textbf{CPU Architecture}    \\ \midrule
    \MA1           & AMD Rome                     \\
    \MA2           & Intel Sandy~/~Ivy Bridge     \\
    \MA3           & Intel Haswell~/~Broadwell    \\
    \MA4           & Intel Skylake~/~Cascade Lake \\ \bottomrule
  \end{tabular}
\end{table}

\paragraph{Success}
Our overall success rates were \SI{70.5}{\percent} for FMNIST, and \SI{28.25}{\percent} for CIFAR10.
\Cref{tbl:ratios} breaks down the successful terminations by the \MAs{} they can identify.
This confirms that an untargeted algorithm is sufficient to generate a set of boundary samples uniquely identifying all \MAs{}, if one runs it repeatedly on different input samples until a matching boundary sample is produced.
Even in the worst case, when identifying \MA4 with FMNIST, a suitable boundary sample is found with more than 99\,\% probability after 28 successful runs, or 40 runs if one accounts for the failure rate of $29.5$\,\%. (Values for CIFAR10 are 24 and 85, respectively.)

\begin{table}[h]
  \caption{Distribution of identified microarchitectures (in percent). \MA{} numbers are scaled to number of successes.}
  \label{tbl:ratios}
  \begin{tabular}{@{}lccccc@{}}
    \toprule
    \textbf{Model}                    & \multicolumn{4}{c}{\textbf{Success}} & \textbf{Failure}                   \\ \midrule
                                      & MA1                                  & MA2              & MA3   & MA4   & \\ \cmidrule(l){2-5}
    \multirow{2}{*}{\textbf{FMNIST}}  & \multicolumn{4}{c}{\textbf{70.50}}   & \textbf{29.50}                     \\
                                      & 29.54                                & 28.47            & 26.33 & 15.66 & \\
    \multirow{2}{*}{\textbf{CIFAR10}} & \multicolumn{4}{c}{\textbf{28.25}}   & \textbf{71.75}                     \\
                                      & 21.24                                & 28.32            & 32.74 & 17.7  & \\ \bottomrule
  \end{tabular}
\end{table}

\paragraph{Transparency}
\Cref{tbl:psnr} reports the distribution of PSNRs for successful boundary samples broken down by dataset.
The PSNR values for most samples from both datasets are above \SI{45}{dB} and thus in the range of high-quality lossy image compression~\cite{welstead1999imagecompression}.
This demonstrates that it is possible to generate a fully identifying set of boundary samples whose elements are not obviously distinguishable from natural samples by a human observer.
By choosing the most transparent boundary samples from the experiments, we could compose fully identifying sets with a minimum PSNR as high as \SI{70.93}{dB} for FMNIST and \SI{77.05}{dB} for CIFAR10.

Boundary samples for CIFAR10 exhibit on average \SI{10}{dB} higher PSNR (and thus better transparency) than FMNIST.
While we cannot causally explain this, it might be related to the larger input dimension which gives more room to distribute the perturbations over more pixels.
Another co-factor is the systematic difference in search time, which we discuss next.

\begin{table}[h]
  \caption{PSNR distribution of boundary samples (in dB).}
  \label{tbl:psnr}
  \begin{tabular}{l cccccc}
    \toprule
    \textbf{Model}   & \textbf{Min} & \textbf{$\bm{Q_1}$} & \textbf{Median} & \textbf{Mean} & \textbf{$\bm{Q_3}$} & \textbf{Max} \\ \midrule
    \textbf{FMNIST}  & \num{38.71}  & \num{46.37}         & \num{50.10}     & \num{52.11}   & \num{55.75}         & \num{83.49}  \\
    \textbf{CIFAR10} & \num{48.28}  & \num{54.76}         & \num{85.76}     & \num{60.82}   & \num{66.65}         & \num{81.48}  \\ \bottomrule
  \end{tabular}
\end{table}

\begin{figure}[h]
  \begin{tikzpicture}[y=.1mm]
    \input{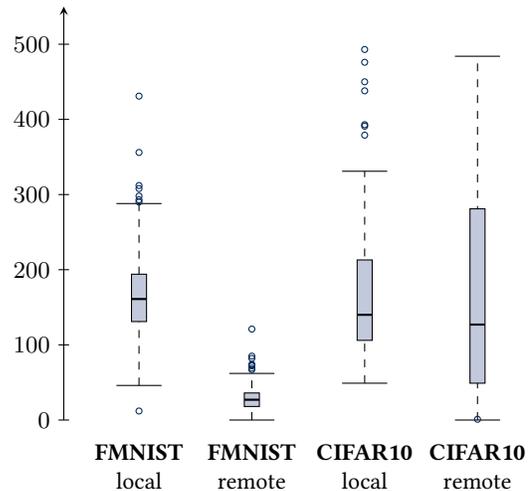}
  \end{tikzpicture}
  \caption{Distribution of the number of local and remote steps across samples.}
  \label{fig:step-distribution}
\end{figure}

\paragraph{Complexity}
\Cref{fig:step-distribution} shows box plots of the distribution of the number of local and remote steps for both datasets.
Both datasets require a few hundred local steps,\footnote{The plot does not show one outlier with 1200 local steps for FMNIST.} but the number of subsequent remote steps varies substantially.
While a few dozen iterations are sufficient for FMNIST, the number of remote steps varies widely for CIFAR10.
Overall, 15.3\% of the successful boundary samples for CIFAR10 required more remote steps than local steps, whereas this happened only once for FMNIST.

\begin{figure}[t]
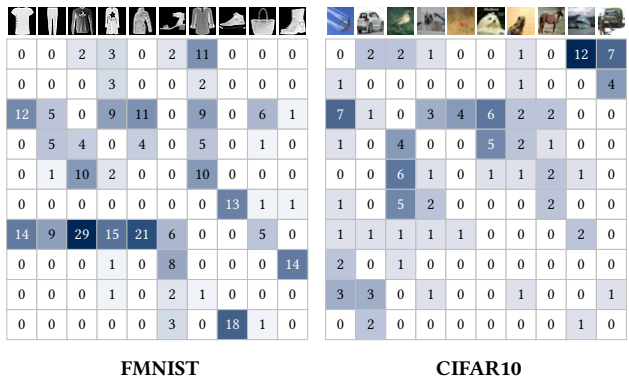

  \begin{subfigure}{0.5\columnwidth}
    \begin{tikzpicture}
      \input{macros/fmnist-confusion.tex}

    \end{tikzpicture}
    \caption*{FMNIST}
    \label{fig:fmnist-confusion}
  \end{subfigure}%
  \begin{subfigure}{0.5\columnwidth}
    \begin{tikzpicture}
      \input{macros/cifar-confusion.tex}
    \end{tikzpicture}
    \caption*{CIFAR10}
    \label{fig:cifar-confusion}
  \end{subfigure}
  \caption{Identifying (rows) and contrast labels (colums). Cell values indicate the frequency of cooccurrence. }
  \label{fig:confusion}
\end{figure}

\paragraph{Favorable class pairs}
Drilling down to the level of class labels, we ask if certain labels are prevalent as identifying or contrast labels.
Recall that the identifying label is the label assigned to the \MA{} singled out by a boundary sample.
Likewise, the contrast label is the one assigned to all other \MAs{}.
\Cref{fig:confusion} depicts confusion matrices with identifying labels in rows and contrast labels in columns.

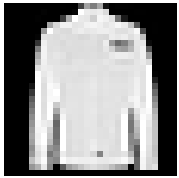
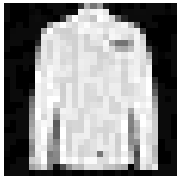
\begin{figure}[h]
  \input{macros/fmnist-sample.tex}
  \caption{FMNIST boundary sample visualization. Perturbations are amplified by a factor of 50 to aid visual inspection.}
  \label{fig:fmnist-sample}
\end{figure}

For FMNIST, the most common identifying label is \emph{Shirt}, and the most common contrasting labels are \emph{T-shirt/Top}, \emph{Pullover}, \emph{Dress}, and \emph{Coat}, which are all closely related in shape.
For CIFAR, the dominance is less pronounced, with the most common label flips being from \emph{Airplane} to \emph{Boat} and \emph{Truck}, which are again of a similar general shape.
The visual similarity of pairs of boundary sample classes can potentially be attributed to the high difficulty of finding boundary samples compared to adversarial samples.
\Cref{fig:fmnist-sample} shows a comparison between natural and boundary sample, including the predicted labels and prediction confidences.

\paragraph{Multivariate analysis}

\begin{figure}[t]
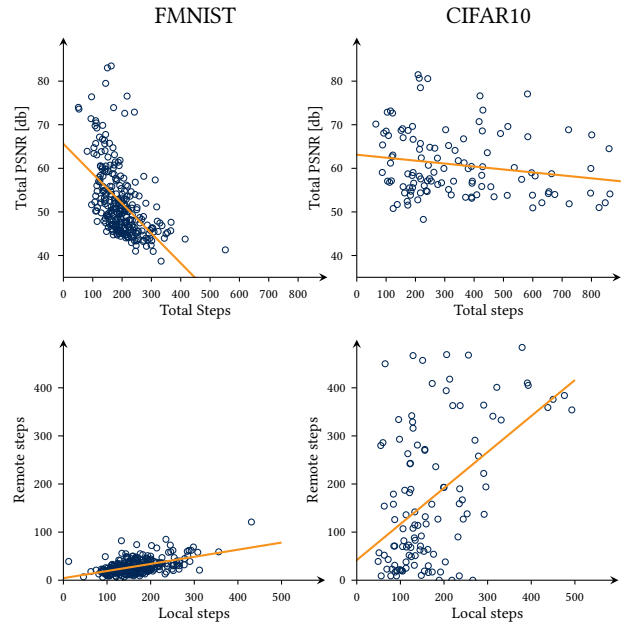

  \centering
  \begin{tikzpicture}
    \providecommand{\psnrx}{.039mm}
    \providecommand{\psnry}{.58mm}
    \providecommand{\stepx}{.058mm}
    \providecommand{\stepy}{.064mm}

    \begin{scope}[x=\psnrx,y=\psnry]
        \begin{scope}
            \clip(0,35) rectangle (900,90);
            \input{macros/fmnist-psnr-steps.tex}
            \draw[uibkorange,thick] (-500, 99.83) -- (900, 4.02);
        \end{scope}

        \draw[->,thin] (0,35) -- (900,35) node[midway, below=7pt] {\scriptsize Total Steps};
        \foreach \x in {0,100,...,800}
        \draw (\x,35) -- ++(0,-1) node[below] {\tiny \x};
        \draw[->,thin] (0,35) -- (0,90) coordinate[midway,left=14.5pt] (psnr);
        \node[rotate=90] at ($(psnr) - (10,0)$) {\scriptsize Total PSNR [db]};
        \foreach \y in {40,50,...,80}
        \draw (0,\y) -- ++(-2pt,0) node[left] {\tiny \y};
        \draw (450, 95) node {FMNIST};
    \end{scope}

    \begin{scope}[xshift=3.9cm,x=\psnrx,y=\psnry]
        \begin{scope}
            \clip(0,35) rectangle (900,90);
            \input{macros/cifar-psnr-steps.tex}
            \draw[uibkorange, thick] (-500,66.5) -- (900,57.03);
        \end{scope}

        \draw[->,thin] (0,35) -- (900,35) node[midway, below=7pt] {\scriptsize Total steps};
        \foreach \x in {0,100,...,800}
        \draw (\x,35) -- ++(0,-1) node[below] {\tiny \x};
        \draw[->,thin] (0,35) -- (0,90) coordinate[midway,left=14.5pt] (psnr);
        \node[rotate=90] at ($(psnr) - (10,0)$) {\scriptsize Total PSNR [db]};
        \foreach \y in {40,50,...,80}
        \draw (0,\y) -- ++(-2pt,0) node[left] {\tiny \y};
        \draw (450, 95) node {CIFAR10};
    \end{scope}

    \begin{scope}[yshift=-2cm,x=\stepx,y=\stepy]
        \begin{scope}
            \clip(0,0) rectangle (600, 550);
            \input{macros/fmnist-remote-local.tex}
            \draw[uibkorange, thick] (-500, -69.62) -- (500, 78.22);
        \end{scope}

        \draw[->,thin] (0,0) -- (600,0) node[midway, below=7pt] {\scriptsize Local steps};
        \foreach \x in {0,100,...,500}
        \draw (\x,0) -- ++(0,-9) node[below] {\tiny \x};
        \draw[->,thin] (0,0) -- (0,500) coordinate[midway,left=14.5pt] (remote);
        \node[rotate=90] at ($(remote) - (10,0)$) {\scriptsize Remote steps};
        \foreach \y in {0,100,...,400}
        \draw (0,\y) -- ++(-2pt,0) node[left=-1pt] {\tiny \y};
    \end{scope}

    \begin{scope}[xshift=3.9cm, yshift=-2cm,x=\stepx,y=\stepy]
        \begin{scope}
            \clip(0,0) rectangle (600, 550);
            \input{macros/cifar-remote-local.tex}
            \draw[uibkorange, thick] (-500, -332.47) -- (500, 416.13);
        \end{scope}

        \draw[->,thin] (0,0) -- (600,0) node[midway, below=7pt] {\scriptsize Local steps};
        \foreach \x in {0,100,...,500}
        \draw (\x,0) -- ++(0,-9) node[below] {\tiny \x};
        \draw[->,thin] (0,0) -- (0,500) coordinate[midway,left=14.5pt] (remote);
        \node[rotate=90] at ($(remote) - (10,0)$) {\scriptsize Remote steps};
        \foreach \y in {0,100,...,400}
        \draw (0,\y) -- ++(-2pt,0) node[left=-1pt] {\tiny \y};
    \end{scope}
\end{tikzpicture}
  \caption{Relation between transparency and complexity.}
  \label{fig:psnr-steps}
\end{figure}

The scatterplots of successful boundary samples in \Cref{fig:psnr-steps} visualize the relationship between the complexity of finding a suitable boundary sample and the resulting transparency (top panels). The complexity is further broken down into local and remote steps in the bottom row.
Unsurprisingly, longer search implies lower quality as difficult samples need stronger perturbations to reach a class boundary.
This interpretation is supported by the positive association between local and remote steps, indicating that the difficulty inherent to the sample determines the search effort.
However, the relation between iterations and (lower) quality is much more pronounced for FMNIST than for CIFAR10.

\section{Related Work}
\label{sec:related}

Adversarial machine learning has become a vast field in the past couple of years.
As it is tangential to our objective, we refer the reader to the authoritative surveys and taxonomies~\cite{papernot2018securityML,akhtarThreat2018}.
In their terminology, boundary samples would represent attacks against supervised machine learning models performed by an iterative evasion during the inference phase.
Another relation is that boundary samples can serve as oracles in gray-box scenarios.

The (still smaller) literature on watermarking neural networks can be broadly structured along two purposes.
First, to protect trained models against unauthorized redistributions;
second, to re-identify models in a black-box or gray-box scenario~\cite{papernot2018securityML},
where only parts of the pipeline are known and accessible.
Our work is closer to the latter, but assumes full knowledge of the model and seeks to identify the execution environment in which it is run.

Uchida et al.~\cite{uchida2017embeddingwatermarks} propose a framework for embedding watermarks into neural networks.
Their method promises to generate a unique signature of the model by adjusting weights in the training phase utilizing a parameter regularizer.
Recent work takes advantage of backdoors inserted in the training phase to activate a detection mechanism with special inputs~\cite{adi2018watermarkbackdoor,zhang2018watermarking,li2019blindwatermark}.
Namba et al.~\cite{namba2019exponentialweighting} combine multiple techniques to implement watermarks, which are reportedly more robust against model and query modifications.
A different approach is to embed a watermark into the distribution 
of the data abstraction obtained in different layers~\cite{rouhani2018deepsigns}.
In order to use watermarking in a black-box scenario, Guo et al.~\cite{guo2018watermarkingembedded} proposed to train a secret message mark into the model, which causes misclassification of certain marked inputs, akin adversarial samples.
Shumailov et al.~\cite{shumailov2020taboo} show a method to embed keys into deep neural networks.
Although they focus on defending against adversarial samples, the method could also be of potential use for watermarking.
Merrer et al.~\cite{merrer2020frontierstitching} use adversarial samples to identify the characteristics of the hyperplane of individual models.
Their generation algorithm is similar to ours (and inspired from the same original method~\cite{goodfellow2014Harnessing}), although it has laxer restrictions.

All approaches discussed in the paragraph above use some form of keys, which are embedded in the model during the training phase.
The ease of securing keys embedded in neural networks has been challenged in last year's workshop~\cite{kupek2020keyleakage}.
The approach presented here is keyless. We are not aware of any other work trying to exploit numerical deviations between execution environments.

\section{Discussion and Future Work}
\label{sec:future}

We have proposed algorithms to generate sets of transparent boundary samples that can identify which microarchitecture (\MA) is being used for predictions, based on the output label alone.
An evaluation of 400 samples from two datasets using ResNet instances of two depth results in success rates between \SI{28.25}{\percent} and \SI{70.50}{\percent}.
The successful samples span all four \MAs{} considered.
This means replacing unsuccessful attempts is a valid strategy.
Transparency in terms of PSNR was almost always at least as good as qualities accepted as imperceptible in the literature on lossy image compression (40--50\,dB).
We showed how specifically composed sets can reach PSNRs of \SI{70}{dB} and higher.
The goal of this paper to establish transparent boundary samples has thus been achieved.

Nevertheless, there is much room for future work.
The two datasets in our study exhibited different difficulty in finding boundary samples.
A third dataset, ImageNet~\cite{imagenet} with much larger input space and higher model complexity has not given us a single boundary sample in reasonable time.
We also only covered the closed set identification scenario, where a set of candidate \MAs{} exists.
Finding boundary samples for larger models, input sizes, (partially) unknown models, and without candidate \MAs{} are open problems.

Another knob to turn is improving the algorithm.
The presented version, inspired by FGSM, is simple, shown to be effective for small instances, but in no way optimal.
For example, we do not even consider PSNR as an objective.
One could take inspiration from other algorithms for finding adversarial samples, PGD~\cite{madry0218pgd} and JSMA~\cite{papernot2016jsma}, which constrain the infinity norm or combine two criteria in a so-called saliency map, respectively.
Devising an algorithm that finds boundary samples targeted to a specific \MA{} is another possible direction.

While boundary samples may enable new forms of digital rights management for trained models, more applications would be possible if the resulting boundary samples survived quantization to eight bit.
This and high PSNR are conflicting goals.

This work has explored the low-hanging fruits.
We used tractable models for a handful of accessible \MAs.
This ad-hoc approach reached limits in terms of resolution (microarchitecture refinements fall together), scope (GPUs not considered), and complexity (both very small and very large models and input dimensions).
Better methods can most likely improve on any of these directions.
But where are fundamental limits?
In short, the boundaries of boundary samples are not yet understood.

\begin{acks}
  The authors thank Nora Hofer for her valuable feedback and help in preparing the camera-ready version of this paper.
\end{acks}

\bibliographystyle{ACM-Reference-Format}
\bibliography{sources}


\end{document}